\documentclass[letterpaper, 10 pt, conference]{ieeeconf}  

\IEEEoverridecommandlockouts                              

\usepackage{graphics} 
\usepackage{epsfig} 
\usepackage{amsmath} 
\usepackage{epstopdf}
\usepackage{lipsum,amsmath,multicol}
\usepackage{float}
\usepackage{algorithm}
\usepackage{algpseudocode}
\usepackage{subfigure}
\usepackage[T1]{fontenc}

\title{\LARGE \bf
Model Predictive Control for Autonomous Driving Based on Time Scaled Collision Cone}

\author{Mithun Babu$^1$, Yash Oza$^1$, Arun Kumar Singh$^2$, K. Madhava Krishna$^1$, Shanti Medasani$^3$ 
\thanks{1. RRC, IIIT-Hyderabad, India 2. AUT, Tampere University of Technology, Finland 3. Mathworks Inc. Equal contribution from first two authors. Third author was supported by TUT Postdoc fellowship. 
}
}

\begin{document}

\maketitle
\thispagestyle{empty}
\pagestyle{empty}
\allowdisplaybreaks

\begin{abstract}
In this paper, we present a Model Predictive Control (MPC) framework based on \emph{path velocity decomposition} paradigm for autonomous driving. The  optimization underlying the MPC  has a two layer structure wherein first, an appropriate path is computed for the vehicle followed by the computation  of optimal  forward velocity along it. The very nature of the proposed path velocity decomposition allows for seamless compatibility between the two layers of the optimization.

A key feature of the proposed work is that it offloads most of the responsibility of collision avoidance to velocity optimization layer for which computationally efficient formulations can be derived. In particular, we extend our previously developed concept of \emph{time scaled collision cone} (TSCC) constraints and formulate the forward velocity optimization layer as a convex quadratic programming problem. We perform validation on autonomous driving scenarios wherein proposed MPC repeatedly solves both the optimization layers in receding horizon manner to compute lane change, overtaking and merging maneuvers among multiple dynamic obstacles.

\end{abstract}

\section{Introduction}

Model Predictive Control or MPC  has emerged as a powerful tool for navigation of car like vehicles. Its strength  stems from the fact  that it can incorporate constraints on state and control while minimizing a user defined cost function. For car-like vehicles, the optimization underlying the MPC becomes strongly non-linear and non-convex due to the motion model of the vehicle and collision avoidance constraints. Current state of the art approaches rely on iterative techniques like sequential convex programming \cite{boyd}, \cite{trajopt} to solve the underlying optimization \cite{alonzomora_icra17}, \cite{nonlinear2}. However,  without a good heuristic for initialization and under the strict real time constraints demanded by autonomous driving, these iterative techniques can sometime fail to provide any feasible solution. As a prospective remedy to this, most MPC frameworks rely on the use of so called "warm start" initializations for the underlying optimization. Herein solutions computed at previous instants are used as an initialization for solving the optimization problem at the current instant.  The MPC frameworks proposed in \cite{linear1}, \cite{linear2} follow an alternate methodology. They directly substitute non-linear motion model with approximate linear form, thus significantly reducing the complexity of the optimization.  However, it should be noted that trajectories thus obtained are not guaranteed to be kino-dynamically feasible or collision free.

%

\subsection{Overview of Proposed Approach and Contributions}

\noindent The proposed approach has a similar flavor as that of the concept of \emph{path velocity decomposition} first proposed in \cite{kant_zucker}. However, the technical approach is significantly different. Specifically, in contrast to \cite{kant_zucker}, the proposed approach considers the reactive setting and uses the concept of collision cone \cite{ghose} for modeling dynamic collision avoidance.  Similar to \cite{kant_zucker}, we observe that the collision avoidance maneuvers has two components namely the path taken by the robot and the forward velocity along it. In line with this observation, in this paper, we propose an MPC framework wherein the underlying optimization has a two layer structure for separate path and  forward velocity optimization. It is essential that both the optimization layers are compatible with  each other. That is, the paths obtained at the first layer should allow us to easily obtain collision free forward velocities along it. In the proposed work, this compatibility is naturally achieved given that both the optimizations solves the same collision cone constraints but with different set of variables. The path optimization solves the collision cone constraints with angular velocity as the variable. The forward velocity optimization solves the so called \emph{time scaled collision cone} (TSCC) constraint \cite{cdc13}, \cite{iros14} which is a projection of collision cone constraints along a given path.

A key feature of the proposed work is that it reduces the reliance on using iterative techniques and solving general non-linear optimizations for computing a collision free trajectory. This is achieved by offloading most of the responsibility of exactly satisfying the collision avoidance constraints to the forward velocity optimization layer for which computationally efficient structures can be derived. To be precise, we extend the concept of TSCC beyond \cite{cdc13}, \cite{iros14} and formulate the forward velocity optimization  as a convex quadratic programming problem. Both the optimizations layers are solved repeatedly in receding horizon manner to obtain lane change, overtaking and lane merging scenarios. 

The rest of the paper is organized as follows. Section \ref{rel} presents a review of the related works. Section \ref{notation} briefly describes the main notations and symbols used in the formulation.  Section \ref{path} introduces the path optimization. Section \ref{tscc} reviews the time scaling concepts and formulates the forward velocity optimization. Section \ref{simplify} presents the reformulation of the forward velocity optimization into a convex quadratic programming problem. Section \ref{tangents} describes the method of encompassing tangents through which the collision avoidance between rectangular objects can be treated in the same way as that between disk shaped objects. Section \ref{res} presents the simulation results validating our proposed formulation.

\section{Related Work}\label{rel}
In this section, we first review the existing approaches for motion planning in dynamic environments for non-holonomic  robots in the backdrop of autonomous driving. Subsequently, we review the more niche approaches specifically designed for autonomous driving.

%

\subsection{Motion Planning in Dynamic Environments.}
\noindent Motion planning in dynamic environments has been extensively studied through the lens of velocity obstacle (VO) \cite{vo} which are analogous and independently developed along with collision cone (CC) \cite{ghose}. The strength of (VO) or (CC) concept is that it models dynamic collision avoidance as a first order constraint in combined position-velocity space. As a result, imminent collisions till some future time instants can be modeled based on just current velocity and positions. This feature has been exploited in works like \cite{rvo1}, \cite{borca}, \cite{gvo} to formulate a one step MPC  framework. Both the concept of VO and CC were originally developed for holonomic robots moving along straight line trajectories. Their extension to non-holonomic robots is non-trivial and still an active area of research. The extension proposed in \cite{borca} is built on the idea that  a non-holonomic robot would exhibit a finite tracking error while tracking a holonomic (piece-wise straight line) trajectory. Moreover, these tracking errors can be modeled as a function of the current state and the commanded holonomic velocities. In other words, this abstraction essentially signifies that for a given tracking error, it is possible to define a space of allowed holonomic velocities. Thus, \cite{borca} proceeds by first  enlarging the radius of the robot by the tracking error that would be observed while simultaneously constraining the collision free velocity to lie within a specific set of allowed holonomic velocities. This last step of constraining the collision free velocity is the key as the set of allowed holonomic velocities are usually non-convex. A convex approximation can be obtained at the expense of unearthing a  reduced space of collision free velocities. 

An alternative extension has been proposed in \cite{gvo} where the VO constraints for non-holonomic robots are solved by exhaustive control sampling. Various heuristics and off line pre-computations are exploited to make the  process computationally efficient. It is clear the effectiveness of this approach depends on the number of control samples which in turn depends on the resolution of discretization. However, it is worth pointing out that the computational efficiency and the resolution of discretization are conflicting criteria in this methodology.


%

The above discussed formulations were primarily designed for general robotic environments with mobile robots. Consequently, their extension to autonomous driving setting poses some key challenges. For example, the technique of enlarging the size of the robot as proposed in \cite{borca} is not suitable for autonomous driving setting where lanes of the road provide a restricted space for maneuvers. Moreover, such size increment was proposed for only disk shaped robots and their extension to rectangular car-like vehicles is not straightforward. In contrast, the control sampling approach of \cite{gvo} is good prospective solution provided good heuristics are available for restriction of search space. A recent approach called control obstacles (CO) \cite{co} was used for computing collision free trajectories in autonomous driving setting \cite{autonovi}. Herein, the robot's kinematics as well as  constraints are linearized to allow for quick computation. However, as mentioned earlier, the trajectories computed through linearized models are not guaranteed to be collision free. The proposed formulation provides some key advantages over these cited works. one such advantage being that it can work with rectangular geometries of the vehicle and can satisfy the collision avoidance constraints for provided non-holonomic motion model.

\subsection{Motion Planning for Autonomous Driving}
\noindent Both sampling/graph search based planners and optimization based approaches are currently used for motion planning for autonomous driving. Either of these approaches can be used in a receding horizon manner to develop a MPC framework. Refer \cite{survey} for an extensive survey on different existing approaches. In the context of the proposed work, we will primarily focus on reviewing the optimization based approaches. Works like \cite{trajopt_car1}, \cite{trajopt_car2}, \cite{ijrr_auto} uses optimization to refine a coarse collision free trajectory obtained from the sampling based planners. A similar approach is presented in \cite{fernet} wherein, the vehicle's motion model is first  approximated as a series of integrators and subsequently optimized trajectories are computed for a set of sampled points. These cited approaches do not explicitly include collision avoidance constraints directly within the optimization. A more holistic approach was presented in \cite{alonzomora_icra17} wherein various kino-dynamic and collision avoidance constraints are directly included in the optimization framework. The proposed formulation shares the same holistic nature but differs from \cite{alonzomora_icra17} in the technical approach followed. As mentioned earlier, we rely on exploiting structure of \emph{path velocity decomposition} to limit our reliance on general non-linear optimization techniques. Moreover, unlike \cite{alonzomora_icra17}, we model collision avoidance between rectangular objects through the concept of encompassing tangents. This is less conservative than the multiple circle approximation used in 
\cite{alonzomora_icra17}.

\section{Symbols and Notations}\label{notation}
We use normal faced lower cased letters to represent scalars while bold faced small case letters would be used to represent vectors or vector valued functions. For example, $\textbf{x}(t) = (x(t), y(t))$ and $\textbf{x}_j(t) = (x_j(t), y_j(t))$  respectively represents the trajectory of the vehicle and the $j^{th}$ obstacle in some future time horizon. With a slight abuse of notation, we will use $\textbf{x}(t_i)$ and $\textbf{x}_j(t_i)$ to represent the respective positions  at time instant $t_i$. Both the vehicle and the obstacles are assumed to be disk shaped with radius $d$ and $d_j$ respectively. The symbols, $v(t_i)$, $\theta(t_i)$ would be used to represent the forward velocity and heading angle of the vehicle at time instant $t_i$. Matrices would be denoted by bold faced higher case letters.

\section{Path Optimization} \label{path} 
Consider the following optimization with variable as the angular velocity $\dot{\theta}(t_i)$ acting in a constant manner over the interval $[t_i \hspace{0.2cm} t_{i+1}]$ and resulting in a trajectory $\textbf{x}(t)$. The forward velocity $v(t_i)$ is kept constant over $[t_i \hspace{0.2cm} t_{i+1}]$. Further, $\Delta t = t_{i+1}-t_i$. 

\small
\begin{subequations}
\begin{align}
\min\limits_{\dot{\theta}(t_i)} J = w_1(\theta(t_{i+1})-\theta_d)^2+w_2\dot{\theta}(t_i)^2\label{cost}\\
(\textbf{x}(t_{i+1}), \dot{\textbf{x}}(t_{i+1}) = \textbf{f}(\textbf{x}(t_i), v(t_i), \theta(t_i))\label{motion_mod}\\
\textbf{c}_{bounds}(v(t_i), \theta(t_i))\leq 0 \label{state_const}
\end{align}
\end{subequations}
\normalsize

\vspace{-0.5cm}

\small
\begin{equation}
\theta_d = atan2(y(t_i)-y_f , x(t_i)-x_f)
\label{theta_desired}
\end{equation}
\normalsize

\vspace{-0.5cm}

\small
\begin{equation}
\textbf{f}(\textbf{x}(t_i), v(t_i), \theta(t_i)) =  \begin{cases}
    x(t_{i+1}) = x(t_i) + \dot{x}(t_{i+1})\Delta t \\
    y(t_{i+1}) = y(t_i) + \dot{y}(t_{i+1})\Delta t \\
    \dot{x}(t_{i+1}) = v(t_i)\cos(\theta(t_i)+\dot{\theta}(t_i)\Delta t)\\
    \dot{y}(t_{i+1}) = v(t_i)\sin(\theta(t_i)+\dot{\theta}(t_i)\Delta t)\\     
  \end{cases}
  \label{nonhol1}
\end{equation}
\normalsize

\vspace{-0.5cm}

\small
\begin{equation}
\textbf{c}_{bounds}(v(t_i), \theta(t_i))\leq 0: \begin{cases}
-\kappa_{max}v(t_i)\leq \dot{\theta}(t_i)\leq \kappa_{max}v(t_i)\\
\vert \dot{\theta}(t_i) \vert \leq \dot{\theta}_{max}
\end{cases}
\end{equation}
\normalsize

\vspace{-0.5cm}

\small
\begin{equation}
\textbf{c}_{avoid}(.)\leq 0:\frac{(\textbf{r}_{j}^T\textbf{v}_{j})^2}{\Vert \textbf{v}_{j}\Vert^2}-\Vert\textbf{r}_{j}\Vert^2+(d_{j}+d)^2\leq 0., \forall j = {1,2..n}
\label{collcone}
\end{equation}
\normalsize
 
 \vspace{-0.5cm}
 
\small 
\begin{equation*}
\textbf{r}_{j}= \begin{bmatrix}
    x({t_{i+1}})-x_j({t_{i+1}})  \\
   y({t_{i+1}})-y_i({t_{i+1}}) \\
\end{bmatrix} , \textbf{v}_{i} =   \begin{bmatrix}
    \dot{x}({t_{i+1}})-\dot{x}_i({t_{i+1}})  \\
   \dot{y}_i({t_{i+1}})-\dot{y}_i({t_{i+1}}) \\
\end{bmatrix}.
\end{equation*}
\normalsize

\noindent The first term in the cost function, (\ref{cost}) is inspired from path following algorithms \cite{purepursuit}, \cite{survey} and aims to align the heading angle of the vehicle to that of the straight line connecting its current position to the goal (refer \ref{theta_desired}). It is easy to see that under such condition, a non-holonomic vehicle can easily navigate towards the goal. The second term of the cost function penalizes the use of high magnitude of angular velocities and induces smoothness in the resulting trajectory. The weights $w_1$, $w_2$ trades-off smoothness with goal reaching behavior. The equality constraints (\ref{motion_mod}) ensures that the position and velocity at time $t_{i+1}$ conform to the motion model of the vehicle (refer \ref{nonhol1}). The inequalities (\ref{state_const}) models the bounds on curvature ($\kappa_{max}$) of the resulting trajectory as well as magnitude of the angular velocity. The inequalities (\ref{collcone}) model the collision avoidance between the vehicle and the dynamic obstacles through the concept of collision cone. Note that we only include such dynamic obstacle in (\ref{collcone}) which are seen as converging from the vehicle's perspective \cite{ghose}.

\subsection{Simplification and Solution Process}
\noindent The main challenge in solving optimization (\ref{cost})-(\ref{collcone}) stems from the non-convex motion model and collision avoidance constraints. Our solution process involves first eliminating the equality constraints (\ref{motion_mod}) by explicitly substituting its right hand side in (\ref{collcone}). We next linearize the resulting collision avoidance constraints to obtain the affine approximation $\textbf{A}_{avoid}\dot{\theta}(t_{i+1})\leq \textbf{B}_{avoid}$. Consequently, we solve a simpler variant of optimization (\ref{cost})-(\ref{collcone}) wherein there are no equality constraints and the collision avoidance constraints have an affine form.

\section{Time Scaled Collision Cone (TSCC)}\label{tscc}
The optimization in the last section provides us with a trajectory $\textbf{x}(t), t\in [t_i \hspace{0.2cm} t_{i+1}]$ which may be very close to being collision free \footnote{From closeness, we mean that the collision avoidance constraints are very close to the boundary of the feasible region i.e $\textbf{c}_{avoid}(.)\approx 0$}. It could also be completely collision free but in general there is no such guarantee since the optimization in the previous section only dealt with an affine approximation of the collision avoidance constraints. In this section, we apply the concept of TSCC over the trajectory $\textbf{x}(t)$ and ensure exact satisfaction of collision avoidance constraints by appropriately modifying the forward velocity profile of the trajectory. To this end, we next present a preview of the time scaling and TSCC concept followed by the construction of the second layer optimization.

\subsection{Time Scaling}

\noindent Given a trajectory, $\textbf{x}(t)$, a change in its time scale from $t$ to $\tau$ does not alter the geometric path associated with the trajectory. In other words, let  $\textbf{x}(t_{i})$ correspond to a position on the trajectory at time $t_{i}$. Then, a change in time scale from $t$ to $\tau$ would mean that the  same point on the trajectory would be reached but now at a different time instant $\tau_i$. It is clear that this change in time of traversal is simply due to the fact the time scaling transformation changes the forward  velocity along the path associated with  $\textbf{x}(t)$. The exact change in velocity and acceleration are characterized by the following set of equations.

\small
\begin{equation}
\begin{array}{lll}
\dot{\textbf{x}}(\tau)= \dot{\textbf{x}}(t)\dot{s}(t)\\
\ddot{\textbf{x}}(\tau) = \ddot{\textbf{x}}(t)(\dot{s}(t))^2+\dot{\textbf{x}}(t)\ddot{s}(t)\\
\frac{dt}{d\tau} = \dot{s}(t)
\end{array}
\label{timescaledef}
\end{equation}
\normalsize

\noindent The term $\frac{dt}{d\tau}$ in (\ref{timescaledef}) is called the scaling function and it decides the transformation between the time scales $t$ and $\tau$.

\noindent Now, assuming $\frac{dt}{d\tau}$ as a linear function in arbitrary time interval $[t_i \hspace{0.2cm} t_{i+1}]$, we obtain the following relation \cite{hauser}.

\small
\begin{equation}
\tau_{i+1}-\tau_i = \frac{2(t_{i+1}-t_i)}{{\dot{s}({t_i})+\dot{s}({t_{i+1}}})}  =  \frac{2\Delta t}{{\dot{s}({t_i})+\dot{s}({t_{i+1}}})} 
\label{timetrans2}
\end{equation}
\normalsize

\noindent Equation (\ref{timetrans2}) describes how an arbitrary time interval $[t_i \hspace{0.2cm} t_{i+1}]$ in the current time scale gets modified to an interval $(\tau_i \hspace{0.2cm} \tau_{i+1})$ in the new time scale $\tau$. 

\noindent Using (\ref{timetrans2}), we can  derive the following expression for $\frac{d^2t}{d\tau^2}$ \cite{hauser}.

\small
\begin{equation}
\frac{d^2t}{d\tau^2} \approx \frac{\dot{s}({t_{i+1}})^2-\dot{s}({t_{i}})^2}{2\Delta t}
\label{uddot} 
\end{equation}
\normalsize

\noindent Equation (\ref{uddot}) suggests that $\frac{d^2t}{d\tau^2}$ is approximated as a constant in time interval $(t_i \hspace{0.2cm} t_{i+1})$ (or alternately $(\tau_i \hspace{0.2cm} \tau_{i+1})$ ), with its value dependent on the magnitude of the scaling function $\frac{dt}{d\tau}$ at the two end of the interval $[t_i \hspace{0.2cm} t_{i+1}]$.

\subsection{TSCC}
\noindent  Using (\ref{timescaledef}) and the discussions presented in the previous section, the space of velocities that can be achieved at position $
{\textbf{x}}(t_{i+1})$ for the vehicle can be characterized in the following manner in terms of the scaling function.

\small
\begin{equation}
\dot{\textbf{x}}(\tau_{i+1}) = \dot{s}({t_{i+1})}\dot{\textbf{x}}(t_{i+1})
\label{collspace}
\end{equation}
\normalsize

\noindent Using (\ref{collspace}), the following time scaled variant of the collision avoidance constraints (\ref{collcone}) can be obtained

\small
\begin{equation}
\frac{((\textbf{r}^s_{ij})^T\textbf{v}^s_{j})^2}{\Vert \textbf{v}^s_{j}\Vert^2}-\Vert\textbf{r}^s_{j}\Vert^2+(d+d_{j})^2\leq 0.
\label{scalecollcone}
\end{equation}
\normalsize

\small 
\begin{equation}
\textbf{r}^s_{ij}= \begin{bmatrix}
    x({t_{i+1}})-x_j({t_{i+1}})  \\
   y({t_{i+1}})-y_j({t_{i+1}}) \\
\end{bmatrix} , \textbf{v}_{i} =   \begin{bmatrix}
    \dot{s}(t_{i+1})\dot{x}({t_{i+1}})-\dot{x}_j(t_{i+1})  \\
  \dot{s}(t_{i+1})\dot{y}({t_{i+1}})-\dot{y}_j(t_{i+1})  \\
\end{bmatrix}.
\end{equation}
\normalsize

\noindent In (\ref{scalecollcone}), the entities $x(t_{i+1}), \dot{x}(t_{i+1}), y(t_{i+1}), \dot{y}(t_{i+1}), \dot{x}_i(t_{i+1}).. $  etc. are known since the vehicle trajectory, $\textbf{x}(t)$ in old time scale is given. Inequalities (\ref{scalecollcone}) can be represented in a compact form as the following generic quadratic inequalities.

\small
\begin{equation}
a_{j}\dot{s}({t_{i+1}})^2+b_{j}\dot{s}({t_{i+1}})+c_{j}\leq 0
\label{quadineq}
\end{equation}
\normalsize

\noindent Where, $a_{j}$, $b_{j}$ and $c_{j}$ are functions of $x(t_{i+1}), \dot{x}(t_{i+1}), y(t_{i+1}), \dot{y}(t_{i+1})... $ etc.

\subsection{Velocity Optimization}
\noindent We now formulate an optimization problem to compute the scaling function $\dot{s}(t), t \in [t_{i}\hspace{0.2cm} t_{i+1}]$ and consequently a collision free forward velocity along the path associated with $\textbf{x}(t)$

\small
\begin{subequations}
\begin{align}
\arg \min_{\dot{s}(t)} J_2 = \sum_{i}(\dot{s}_i({t_{i+1}})-\dot{s}^{pref})^.\label{opt3}\\
v^{min}\leq \dot{s}(t_{i+1})\sqrt{ \dot{x}({t_{i+1}})^2+\dot{y}({t_{i+1}})^2}  \leq v^{max} \label{velbound}.\\
\frac{a^{min}}{\sqrt{2}}\leq\dot{s}(t_{i+1})^2\ddot{x}({t_i}) +\ddot{s}(t_i)\dot{x}(t_i)\leq \frac{a_{i}^{max}}{\sqrt{2}} .\label{accbound1}\\
\frac{a^{min}}{\sqrt{2}}\leq\dot{s}(t_{i+1})^2\ddot{y}({t_i}) +\ddot{s}(t_i)\dot{y}(t_i)\leq \frac{a_{i}^{max}}{\sqrt{2}} .\label{accbound2}\\
a_{j}\dot{s}({t_{i+1}})^2+b_{j}\dot{s}({t_{i+1}})+c_{j}\leq 0, \forall j={1,2,3..n} .\label{quadine2}\\
s({t_i}) = 1, .\label{continuity} \forall i = 1,2,3....n
\end{align}
\end{subequations}
\normalsize
 
\small
\begin{equation}
\dot{s}^{pref} = \frac{v^{pref}}{\sqrt{ \dot{x}_i({t_{i+1}})^2+\dot{y}({t_{i+1}})^2}}
\label{vref}
\end{equation}
\normalsize

\noindent The cost function in (\ref{opt3}) seeks to compute such scaling functions which  minimizes the deviation of the vehicle from its preferred forward velocity, $v^{pref}$. This preference velocity calculated based on output of Behavioural layer and average lane speed. The chosen design of the cost function assumes a specially important role in the context of autonomous driving. For example, in overtaking scenarios, $v^{pref}$  should necessarily be higher the dynamic obstacles it is overtaking. Inequalities (\ref{velbound}) enforces the constraints that the forward velocity resulting from time scaling transformation should respect the minimum and maximum  bounds. In our formulation, $v_{min}$ is strictly greater than zero to account for the fact that in autonomous driving scenarios, brining the vehicle to a halt may disrupt the traffic flow. Inequalities (\ref{accbound1})-(\ref{accbound2}) enforces the acceleration bounds. For computational reasons, we have split the acceleration bounds into equivalent bounds for the separate $x$ and $y$ components. Inequalities (\ref{quadine2}) is same as (\ref{quadineq}). Finally, the equality (\ref{continuity}) ensures that the velocity profile resulting from the time scaling transformation has continuity with the current velocity profile at initial time instant $t_i$. 

\section{Simplification Of Velocity Optimization}\label{simplify}

In this section, we show that velocity optimization (\ref{opt3})-(\ref{continuity}) can be reformulated to a convex quadratic programming ({QP}) problem. We start with the convexification of {TSCC} constraints (\ref{scalecollcone}) or (\ref{quadineq}) and then subsequently show that its intersection space with velocity and acceleration bounds can be described  in terms of linear inequalities. For the ease of exposition, we introduce the following change of variables \cite{hauser}.

\small
\begin{equation}
\dot{s}({t_i})^2 = z_i({t_i}), \dot{s}({t_{i+1}})^2 = z_i({t_{i+1}}).
\label{change_variables} 
\end{equation}
\normalsize

\noindent With the help of (\ref{uddot}), we can also obtain the following relation .

\vspace{-0.2cm}

\small
\begin{equation}
\ddot{s} \approx \frac{z(t_{i+1})-z(t_i)}{2\Delta t}.
\label{uddot2} 
\end{equation}
\normalsize

\subsection{Convexification of TSCC Constraints}

\noindent In our past work \cite{cdc13}, \cite{iros14}, we have shown the convexity of (\ref{quadineq}) depends on the sign of $a_j$, which in turn depends on the relative positions and velocities between the vehicle and the dynamic obstacles. There, we exploited the fact that for each dynamic obstacle (\ref{quadineq}) represents a single variable quadratic inequality and thus can be solved in closed form. Subsequently, we employed a sorting algorithm to compute the  resultant solution space. However, we have observed that such process become computationally expensive especially if many non-convex instances of (\ref{quadineq}) are present. Thus, here, we present an alternate methodology wherein two new insights are provided. Firstly, we show that even when $a_j<0$, inequalities (\ref{quadineq}) can be converted to a convex form provided we have $c_j\geq 0$. Secondly, we show that for the non-convex instance of (\ref{quadineq}), we can linearize it to obtain an approximation which is more conservative than the original constraints. In other words, satisfaction of linearized constraint guarantees satisfaction of the original constraints.

The following analysis is for one dynamic obstacle but can be easily extended to the general case of $n$ obstacles

\subsubsection{$a_{j}\geq 0$, $c_{j}\leq 0$}\label{case1}
The solution space of (\ref{quadineq}) for this case is given by

\small
\begin{equation}
\dot{s}(t_{i+1}) \hspace{0.2cm}\epsilon \hspace{0.2cm}[\dot{s}^j_{min} \hspace{0.2cm}\dot{s}^j_{max}] .
\label{cas1minmax}
\end{equation}
\normalsize

\tiny
\begin{equation}
\dot{s}^j_{min} = \min(\frac{-b_{j}+\sqrt{(b^2_{j}-4a_{j}c_{j})}}{2a_{j}} , \frac{-b_{j}-\sqrt{(b^2_{j}-4a_{j}c_{j})}}{2a_{j}}).
\end{equation}
\normalsize

\tiny
\begin{equation}
\dot{s}^j_{max} = \max(\frac{-b_{j}+\sqrt{(b^2_{j}-4a_{j}c_{j})}}{2a_{j}},\frac{-b_{j}-\sqrt{(b^2_{j}-4a_{j}c_{j})}}{2a_{j}}).
\end{equation}
\normalsize

\noindent Using (\ref{change_variables}), we have

\small
\begin{equation}
z({t_{i+1}})\geq (\dot{s}^j_{min})^2, \hspace{0.2cm} z({t_{i+1}})\leq (\dot{s}^j_{max})^2.
\label{final2cas1}
\end{equation}
\normalsize

\subsubsection{$a_{j}\leq 0$, $c_{j}\geq 0$}\label{case2}
In this section, we start with writing (\ref{quadineq}) in a slightly different form as below

\small
\begin{equation}
a_j+b\frac{1}{\dot{s}(t_{i+1})}+c_j\frac{1}{{\dot{s}(t_{i+1})^2}}\leq 0
\end{equation}
\normalsize

\noindent The above inequality is convex with respect to $\frac{1}{\dot{s}(t_{i+1})}$ and whose solution space can be characterized in the following form.

\small
\begin{equation}
\frac{1}{\dot{s}(t_{i+1})} \hspace{0.2cm}\epsilon \hspace{0.2cm}[\dot{s}^j_{min} \hspace{0.2cm}\dot{s}^j_{max}] \Rightarrow \dot{s}(t_{i+1}) \hspace{0.2cm}\epsilon \hspace{0.2cm}[\frac{1}{\dot{s}^j_{max}} \hspace{0.2cm}\frac{1}{\dot{s}^j_{min}}] . 
\label{cas2minmax}
\end{equation}
\normalsize

\tiny
\begin{equation}
 \dot{s}^j_{min} = \min(\frac{-b_{j}+\sqrt{(b^2_{j}-4a_{j}c_{j})}}{2c_{j}},\frac{-b_{j}-\sqrt{(b^2_{j}-4a_{j}c_{j})}}{2c_{j}}).
\end{equation}
\normalsize

\tiny
\begin{equation}
\dot{s}^j_{max} = \max(\frac{-b_{j}+\sqrt{(b^2_{j}-4a_{j}c_{j})}}{2c_{j}},\frac{-b_{j}-\sqrt{(b^2_{j}-4a_{j}c_{j})}}{2c_{j}}).
\end{equation}
\normalsize

\noindent Finally, using (\ref{change_variables}), we have

\small
\begin{equation}
z({t_{i+1}})\geq \frac{1}{(\dot{s}^j_{max})^2}, \hspace{0.2cm} z({t_{i+1}})\leq \frac{1}{(\dot{s}^j_{min})^2}.
\label{final2cas1}
\end{equation}
\normalsize

\subsubsection{$a_j\geq0, c_j\geq0$}
\noindent The previous two discussed cases are a subset of this case and any of the approaches can be followed.

\subsubsection{$a_j\leq0, c_j\leq0, b_j\geq0$}
This represents a  non-convex instance of (\ref{quadineq}) and thus is the most difficult to handle.  We start with recalling (\ref{change_variables}) and writing the (\ref{quadineq}) in the following form

\small
\begin{equation}
a_jz(t_{i+1})+b_j\sqrt{z(t_{i+1})}+c_j\leq 0
\label{convex_concave1}
\end{equation}
\normalsize

\noindent Inequality (\ref{convex_concave1}) is in the so called \emph{convex-concave} form wherein the first term is convex while the second term is purely concave. We linearize the second term around some initial guess $z^*$ resulting in the following affine approximation for (\ref{convex_concave1}).

\small
\begin{equation}
a_jz(t_{i+1}) +b_j(\sqrt{z^*}+\frac{1}{2\sqrt{z^*}}(z(t_{i+1})-z^*))+c_j\leq 0 
\label{affine}
\end{equation}
\normalsize

\noindent An interesting point about (\ref{affine}) is that it is a more conservative constraint than (\ref{convex_concave1}) \cite{boyd}. Thus, satisfaction of (\ref{affine}) guarantees satisfaction of (\ref{convex_concave1}).

\subsection{Velocity and Acceleration Bounds}

\noindent Squaring both sides of (\ref{velbound}), we obtain

\small
\begin{eqnarray}\nonumber
(v^{min})^2\leq\dot{s}({t_{i+1}})^2({x}({t_{i+1}})^2+{y}({t_{i+1}})^2)\leq (v^{max})^2.\\
\Rightarrow (v^{min})^2 \leq z({t_{i+1}})(({x}({t_{i+1}})^2+{y}({t_{i+1}})^2)\leq (v^{max})^2
\label{scalevelineq}
\end{eqnarray}
\normalsize

\noindent Using (\ref{change_variables}), (\ref{uddot2}), the acceleration bounds can be put in the following form.

\small
\begin{eqnarray}
\frac{a^{min}}{\sqrt{2}}\leq z({t_{i}})\ddot{x}(t_{i})+\dot{x}(t_{i})\frac{z({t_{i+1}})-z_i({t_{i}})}{2\Delta t}\leq \frac{a^{max}}{\sqrt{2}}\label{scaledacc2}.\\
\frac{a^{min}}{\sqrt{2}}\leq z({t_{i}})\ddot{y}(t_{i})+\dot{y}(t_{i})\frac{z({t_{i+1}})-z_i({t_{i}})}{2\Delta t}\leq \frac{a^{max}}{\sqrt{2}}\label{scaledacc3}.
\end{eqnarray}
\normalsize

\noindent Clearly, (\ref{scaledacc2})-(\ref{scaledacc3}) are set of linear inequalities in terms of variable $z({t_i}), z({t_{i+1}})$.

\noindent {\bf{Summary}}: The discussions presented in this section has shown that the {TSCC} constraints and the velocity and acceleration bounds can  all be represented as linear inequalities in terms of variable $z({t_i}), z({t_{i+1}})$. Let these  inequalities be compactly represented as $\textbf{A}_{inq}{\textbf{z}}-\textbf{b}_{ineq}\leq 0$, where $\textbf{z}= (z({t_i}), z_(t_{i+1}))$. The optimization (\ref{opt3})-(\ref{continuity}) can then be presented in the following simple form.

\vspace{-0.4cm}

\small
\begin{subequations}
\begin{align}
 \arg \min_{z({t_i}), z(t_{i+1})} J_2 = \sum_{i}(z_i({s_c})-(s_i^{pref})^2)^2.\label{opt4}\\
\textbf{A}_{inq}{\textbf{z}}-\textbf{b}_{ineq}\leq 0\label{reform1}.\\
z_i({t_{i+1}})\geq 0\label{reform2}.\\
z_i({t_i}) = 1.\label{reform3}
\end{align}
\end{subequations}
\normalsize

\noindent As can be seen, (\ref{opt4})-(\ref{reform3}) is a convex
{QP} problem. The variables in the optimization
are $z({t_i}), z({t_{i+1}}) $. The scaling functions can be easily recovered through $\dot{s}({t_i}) =
\sqrt{z({t_i})}, \dot{s}({t_{i+1}}) = \sqrt{z(t_{i+1}})$.

\section{Method of Encompassing Tangents}\label{tangents}
The formulations presented in the last couple of sections have assumed a disk shape for the vehicle and dynamic obstacles. In this section, we show that using the method of encompassing tangents \cite{ghose}, the same formulation can be easily adapted for rectangular shaped objects as well. The following presentation is a slight adaptation of the original concept proposed in \cite{ghose}. 

\begin{figure}[!h]
  \centering
   \subfigure[]{
    \includegraphics[width=8.3 cm, height=1.3 cm]{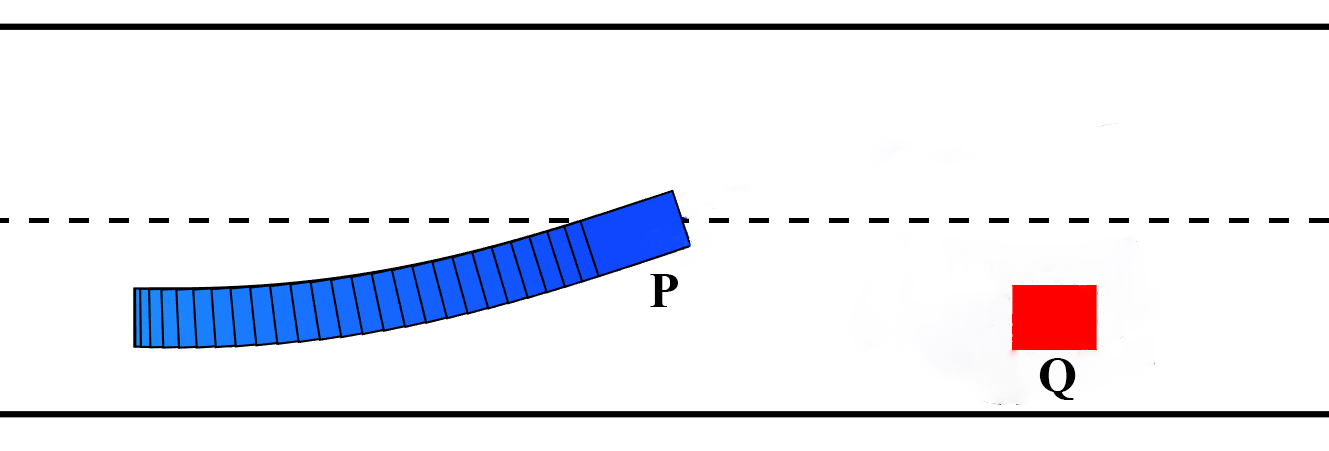}
    \label{onroad_1}
   }\hspace{-0.2cm}
   \subfigure[]{
    \includegraphics[width=8.6 cm, height=2.3 cm]{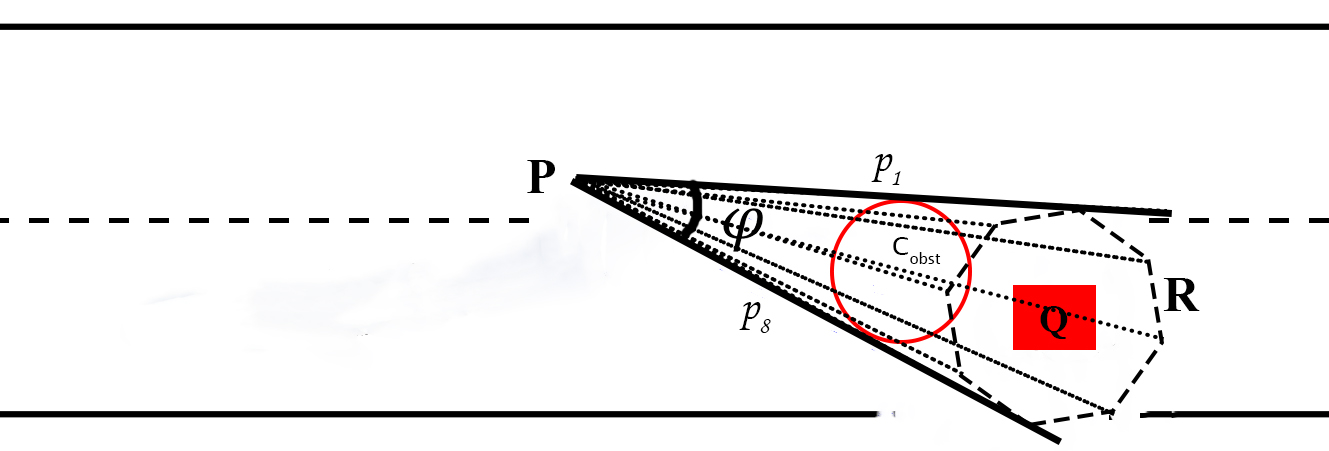}
    \label{onroad_2}
   }
\caption{(a) Collision modeling between rectangular objects. The blue represents the planned vehicle while the red represents the dynamic obstacles. (b) The planned vehicle is reduced to a point and the dynamic obstacle is enlarged through the concept of Minkowski sum. The solid black lines represent the so called encompassing tangents. Collision avoidance between rectangles can be modeled as that between a point and a circle ($C_{obst}$) constructed based on encompassing tangents. }
\end{figure}

Consider Fig.~\ref{onroad_1} where $P$ (vehicle) and $Q$ (dynamic obstacle) represent two rectangular objects. A Minkowski sum boundary polygon $R$, can be mathematically represented as $R= Q\oplus(-P)$, Where $P\oplus Q = [p+q \mid p\in P, q\in Q]$. As shown in Fig.~\ref{onroad_2},  the Minkowski sum allows us to reduce the rectangle $P$ to a point and enlarge $Q$ into polygon $R$. The minkowski sum $R$ can be computed in $O(m+n)$ time for two polygons with m and n edges. Next, we compute the so called encompassing tangents (solid black line in Fig.~\ref{onroad_2}) which are straight lines drawn from $P$ to polygon $R$ with an enclosing angle $\varphi$ given by
$$\varphi = max_{i,j} (\mid \arctan(p_i) - \arctan(p_j)  \mid)$$ where $p_i$ is the slope of the lines drawn from $P$ to $i^{th}$ vertex of $R$. Following \cite{ghose} and as shown in Fig.~\ref{onroad_2}, the collision avoidance between rectangles $P$ and $Q$ can be equivalently modeled as that between a point and the circle $C_{obst}$ constructed on the  basis of encompassing tangents. Their velocities  will be same as that of $P$ and $Q$ respectively. Note that the $C_{obst}$  is much less conservative than the circumscribing circle of $R$. $C_{obst}$ also has less area than that obtained by representing the polygon $R$ with multiple circles \cite{alonzomora_icra17}.

\section{Simulations Results}\label{res}

\subsection{MPC and Simulation Framework}
\noindent The employed MPC framework is simple and consists of repeatedly solving the path and velocity optimization and executing the computed trajectory. Nevertheless, few key points are worth pointing out. Firstly, only a part of the current computed trajectory is executed before a new set of computations are initiated. In other words, our planning horizon is much smaller (typically 10 times) than the planning horizon which is typical for an MPC framework. Secondly, in autonomous driving scenarios the vehicle is required to follow a path rather than reach a specific goal position. In our MPC implementation, we induce this behavior by giving a pseudo velocity to the goal position. The magnitude of this pseudo-velocity is tuned to get the best results. Finally, to constrain the vehicle to stay within the road boundaries, we created some fictitious static obstacles along the boundaries and included them in the path optimization framework as dynamic obstacles with zero velocities. Simulations were performed in Gazebo \cite{gazebo} while optimizations were solved in Python with CvxOpt \cite{cvxopt} as the solver. Gazebo is a 3D simulator developed  and supports multiple physics engines. The obstacle locations and their velocities were sensed by LIDAR mounted on top of car. In the Gazebo snapshots shown later, the gray color SUV represents the  vehicle while rest of the cars are dynamic obstacles. A simulation video can be found at https://www.youtube.com/watch?v=8re0I9z0LEk\& feature=youtu.be

\subsection{Benchmark Scenarios}
\subsubsection{Overtaking}

\begin{figure}[!h]
  \centering
   \subfigure[]{
    \includegraphics[width=2.3 cm, height=4.0 cm]{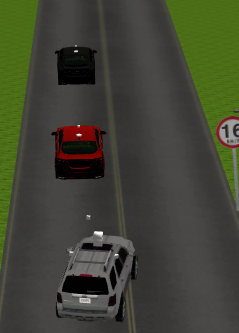}
    \label{fin2_overtaking}
   }\hspace{-0.2cm}
   \subfigure[]{
    \includegraphics[width=2.3 cm, height=4.0 cm]{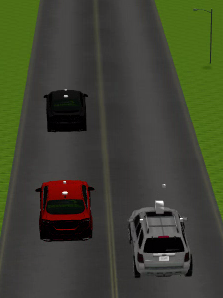}
    \label{fin3_overtaking}
   }
   \subfigure[]{
      \includegraphics[width=2.3 cm, height=4.0 cm]{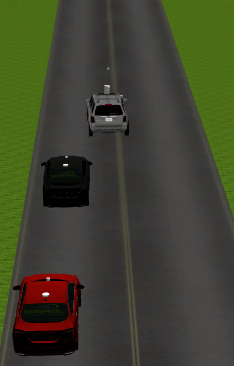}
    \label{fin5_overtaking}
   }
   \subfigure[]{
     \includegraphics[width=8.6 cm, height=1.3 cm]{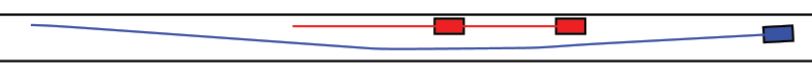}
    \label{overall_overtaking}
   }\vspace{-0.3cm}
   \subfigure[]{
     \includegraphics[width=8.6 cm, height=2.0 cm]{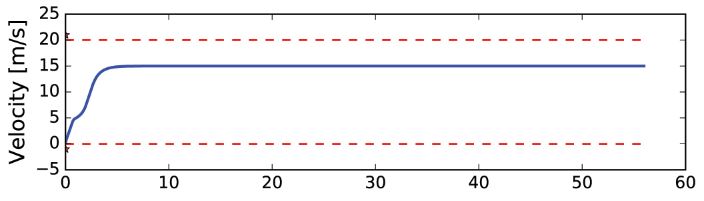}
    \label{vel_overtaking}
   }\vspace{-0.3cm}
   \subfigure[]{
     \includegraphics[width=8.6 cm, height=2.0 cm]{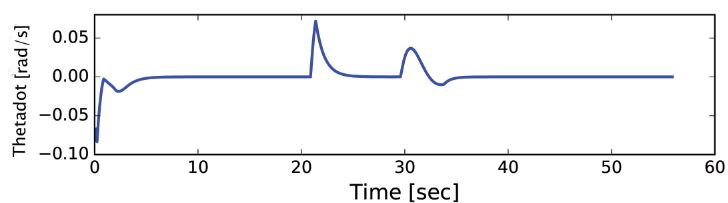}
    \label{omega_overtaking}
   }   
  \caption{ (a)-(c): Gazebo snapshots of overtaking maneuver. (d): Path of the planned vehicle. (e)-(f): Forward and angular velocity plots.}          
\end{figure}

 \noindent Fig.~\ref{fin2_overtaking}-\ref{fin5_overtaking} shows the simulation of overtaking maneuvers in Gazebo. The dynamic obstacles are all moving at a speed of $5m/s$ along straight line paths. Fig.~\ref{overall_overtaking} shows the path of the  concerned vehicle. As can be seen, the vehicle first departs from the lane and comes back to it after overtaking. The forward velocity plots shown in Fig.~\ref{vel_overtaking} reveal an interesting trend.  The initial forward velocity is around $2m/s$ for the vehicle which is significantly less than the dynamic obstacles ahead. Thus, to initiate an overtaking maneuver, we set $v_{pref}=15m/s$ in (\ref{opt3}). As can be seen from Fig.~\ref{vel_overtaking}, the vehicle gradually increases its forward velocity and settles down at $15m/s$ after overtaking. The angular velocity plots are shown in Fig.~\ref{omega_overtaking} and exhibit a  smooth profile.

\subsubsection{Overtaking and Following}

\begin{figure}[!h]
  \centering
   \subfigure[]{
    \includegraphics[width=2.3 cm, height=4.0 cm]{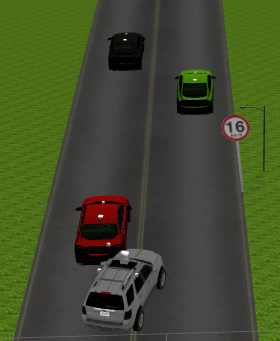}
    \label{outro22}
   }\hspace{-0.2cm}
   \subfigure[]{
    \includegraphics[width=2.3 cm, height=4.0 cm]{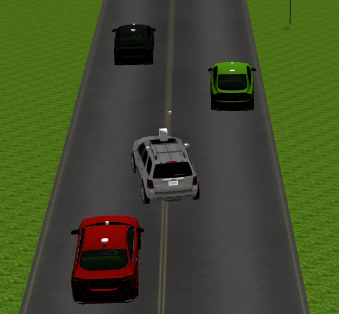}
    \label{outro24}
   }
   \subfigure[]{
      \includegraphics[width=2.3 cm, height=4.0 cm]{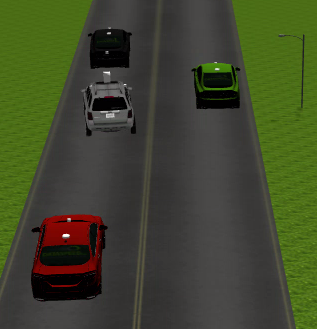}
    \label{outro25}
   }
   \subfigure[]{
     \includegraphics[width=8.6 cm, height=1.0 cm]{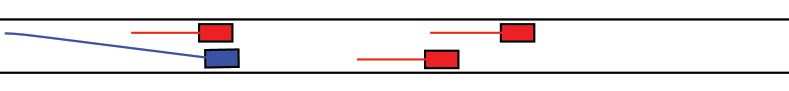}
    \label{overall_1}
   }
   \subfigure[]{
     \includegraphics[width=8.6 cm, height=1.0 cm]{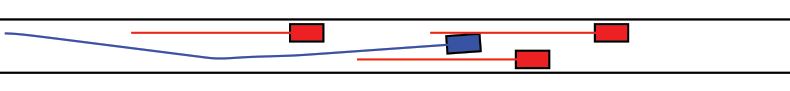}
    \label{overall_2}
   }
   \subfigure[]{
     \includegraphics[width=8.6 cm, height=1.0 cm]{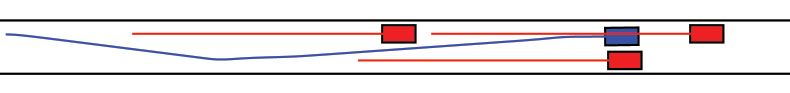}
    \label{overall_3}
   }\vspace{-0.3cm}
   \subfigure[]{
     \includegraphics[width=8.6 cm, height=2.0 cm]{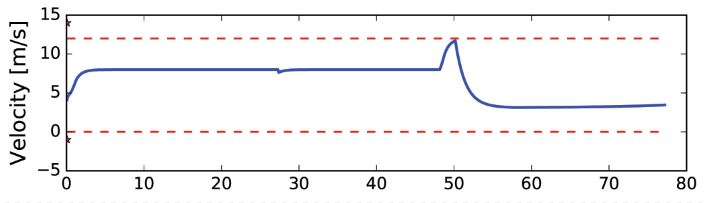}
    \label{vel_compli}
   }\vspace{-0.3cm}
   \subfigure[]{
     \includegraphics[width=8.6 cm, height=2.0 cm]{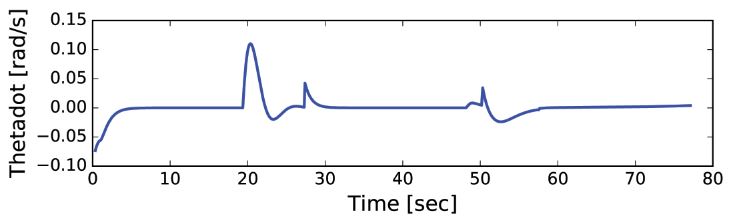}
    \label{omega_compli}
   }   
  \caption{(a)-(c): Gazebo snapshots of overtaking combined with following maneuver. (d)-(f): Path of the planned vehicle at three different time instants. (g)-(h): Plots of forward and angular velocity. }          
\end{figure}

\noindent Fig.~\ref{outro22}-\ref{overall_3} show the  concerned vehicle change lane to overtake a slow moving dynamic obstacle and then slow down to follow the dynamic obstacle ahead. The behavior can also be cross-validated against the forward velocity plots shown in Fig.~\ref{vel_compli}. The forward velocity increases to $v^{pref}=8m/s$ during the initiation of the overtaking and then decreases to $4m/s$ after coming back to the lane. In between, the forward velocity jumps momentarily to $12m/s$ to quickly complete the overtaking maneuver. The angular velocity plots shown in Fig.~\ref{omega_compli} again exhibit a smooth profile.

%
%

\subsubsection{Merging and overtaking}

\begin{figure}[!h]
  \centering
   \subfigure[]{
    \includegraphics[width=2.3 cm, height=4.0 cm]{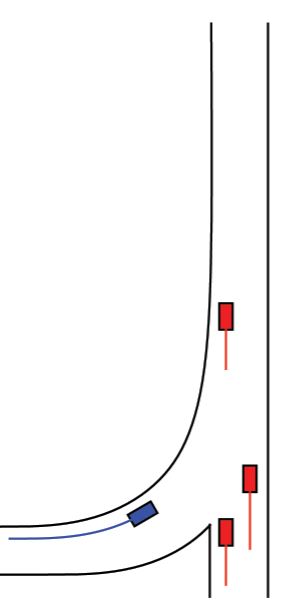}
    \label{merge1}
   }\hspace{-0.2cm}
   \subfigure[]{
    \includegraphics[width=2.3 cm, height=4.0 cm]{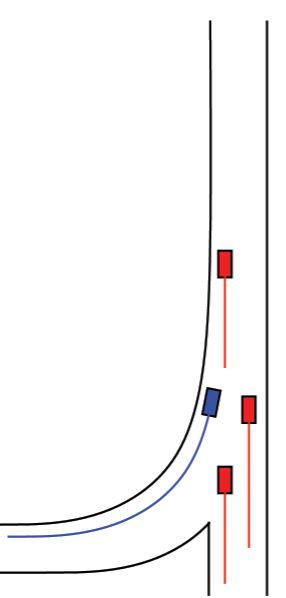}
    \label{merge2}
   }
   \subfigure[]{
      \includegraphics[width=2.3 cm, height=4.0 cm]{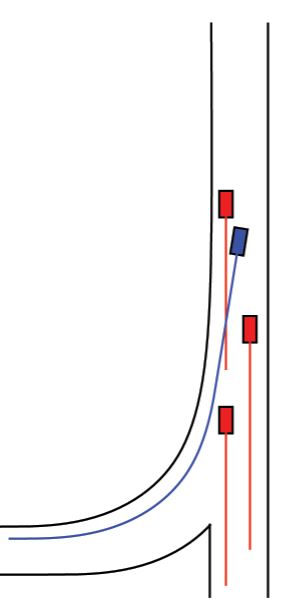}
    \label{merge3}
   }
   \subfigure[]{
     \includegraphics[width=2.3 cm, height=4.0 cm]{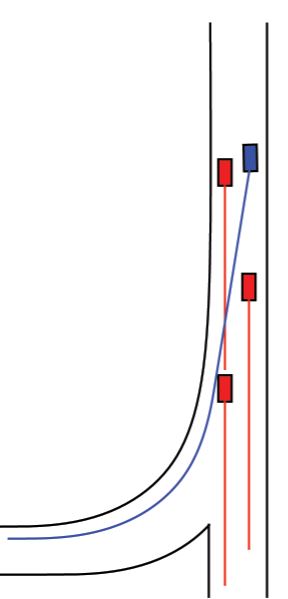}
    \label{merge4}
   }
   \subfigure[]{
     \includegraphics[width=2.3 cm, height=4.0 cm]{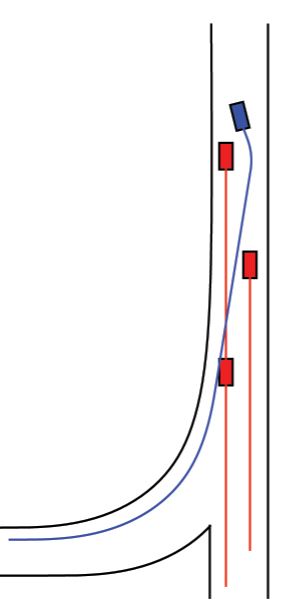}
    \label{merge_5}
   }
   \subfigure[]{
     \includegraphics[width=2.6 cm, height=4.0 cm]{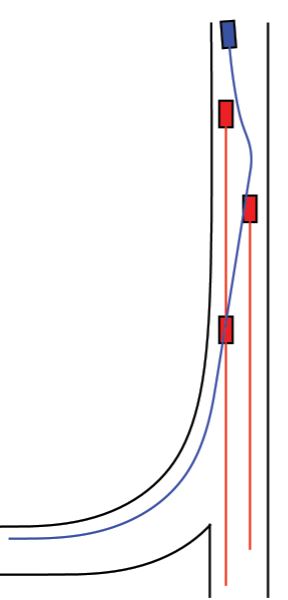}
    \label{merge6}
   }\vspace{-0.3cm}
   \subfigure[]{
     \includegraphics[width=8.6 cm, height=2.0 cm]{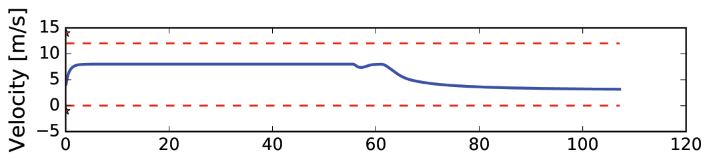}
    \label{vel_merge}
   }\vspace{-0.3cm}
   \subfigure[]{
     \includegraphics[width=8.6 cm, height=2.0 cm]{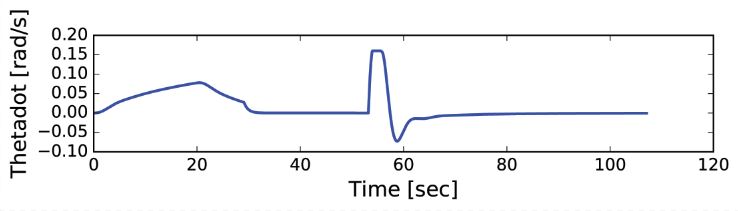}
    \label{omega_merge}
   }   
  \caption{ (a)-(f): Snapshots and path of the planned vehicle for lane merging followed by overtaking maneuver. (g)-(h): Plots of forward and angular velocity.}          
\end{figure}

\noindent Fig.~\ref{merge1}-\ref{merge6} show the concerned vehicle first moving along a free lane and then converging to a lane with slow moving dynamic obstacles followed by an overtaking maneuver. As shown in Fig.~\ref{vel_merge}, the forward velocity increases to the preferred magnitude of $
 8m/s$ and then slows down to $4m/s$ to converge and overtake. The angular velocity plots are shown in Fig.~\ref{omega_overtaking} and is similar in nature to that obtained for the previous cases.

\subsubsection{Overtaking and Lane Change}
\begin{figure}[!h]
  \centering
   \subfigure[]{
    \includegraphics[width=8.6 cm, height=1.0 cm]{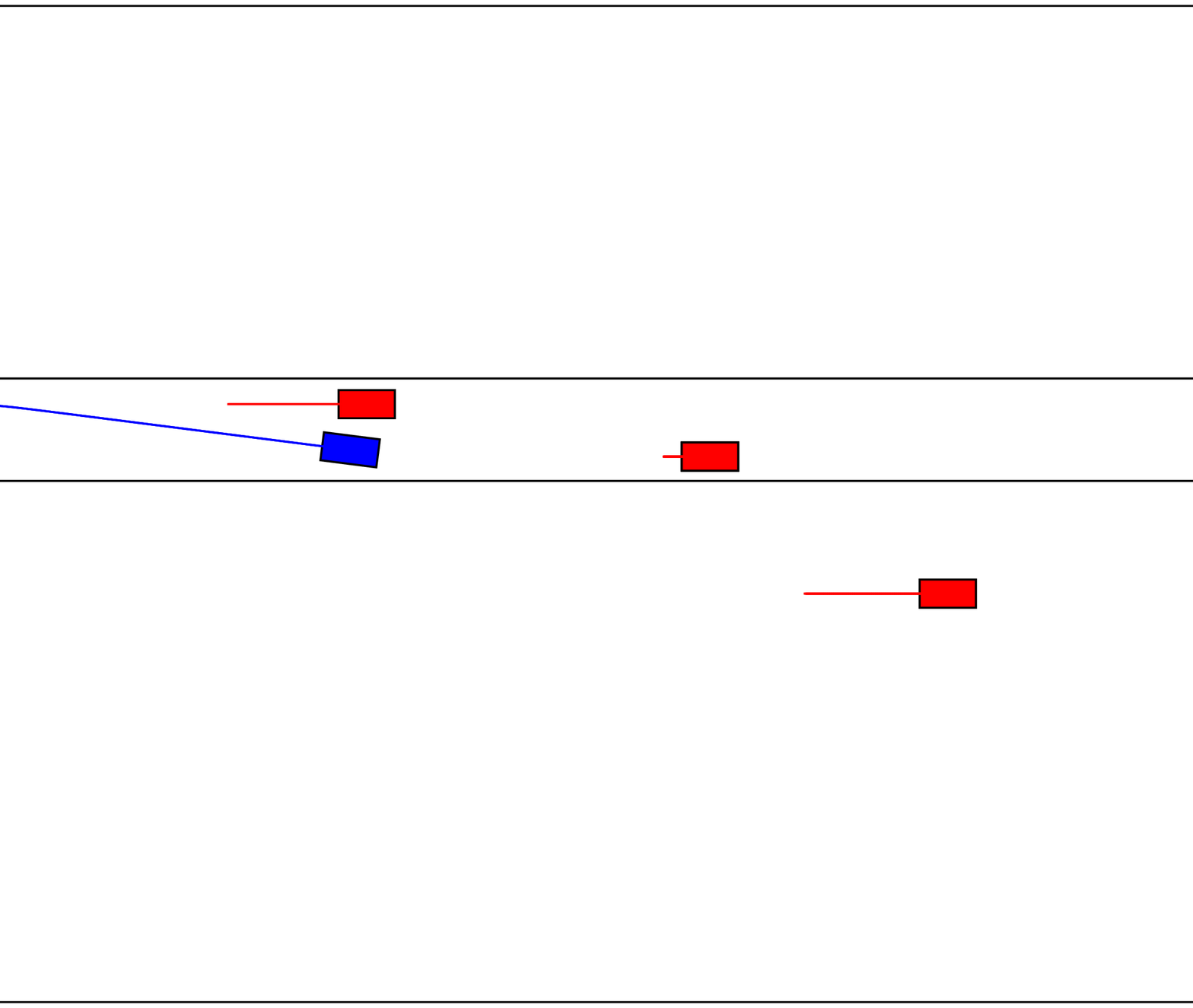}
    \label{lane_1}
   }\vspace{-0.2cm}
   \subfigure[]{
    \includegraphics[width=8.6 cm, height=1.0 cm]{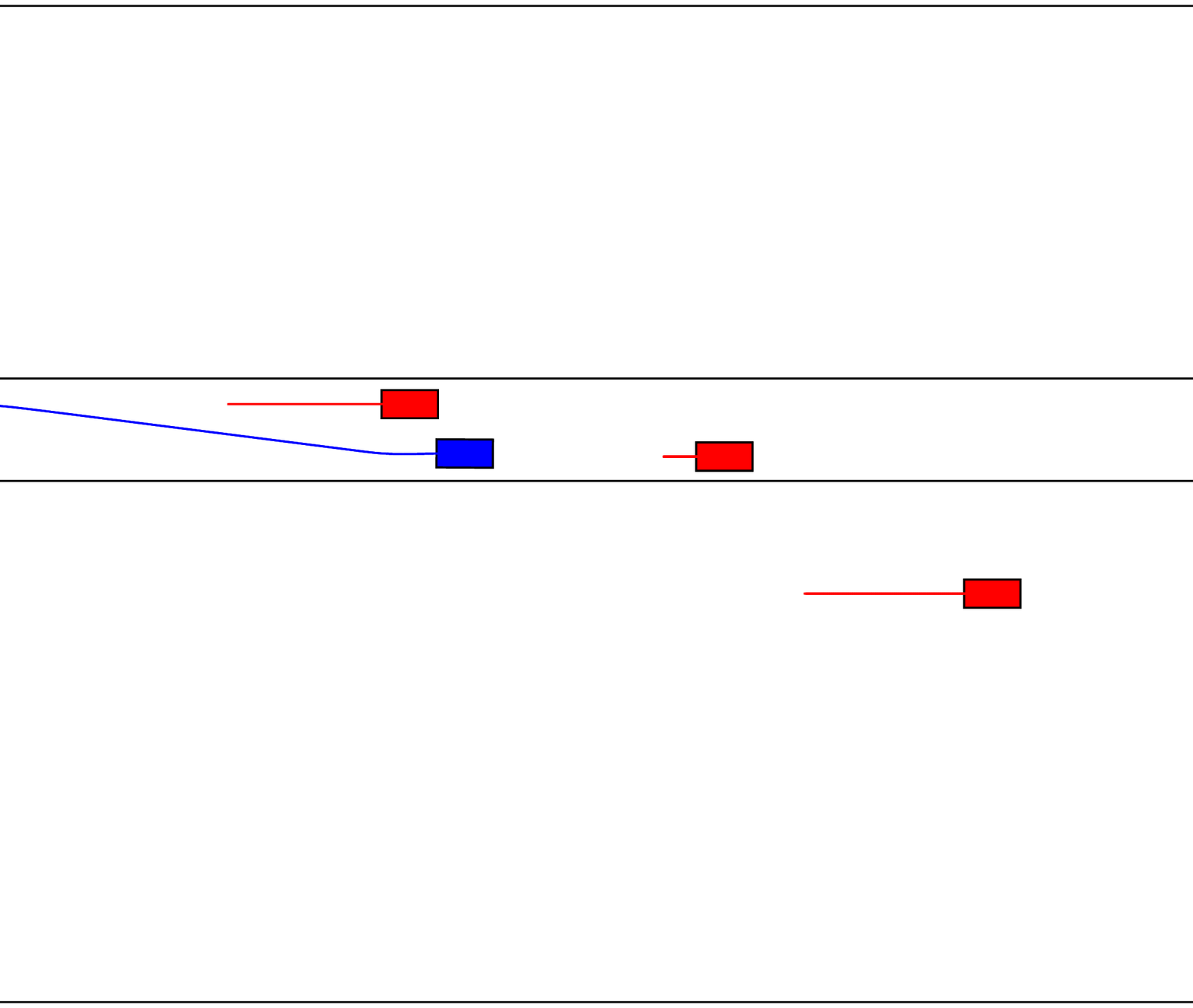}
    \label{lane_2}
   }\vspace{-0.2cm}
   \subfigure[]{
    \includegraphics[width=8.6 cm, height=1.0 cm]{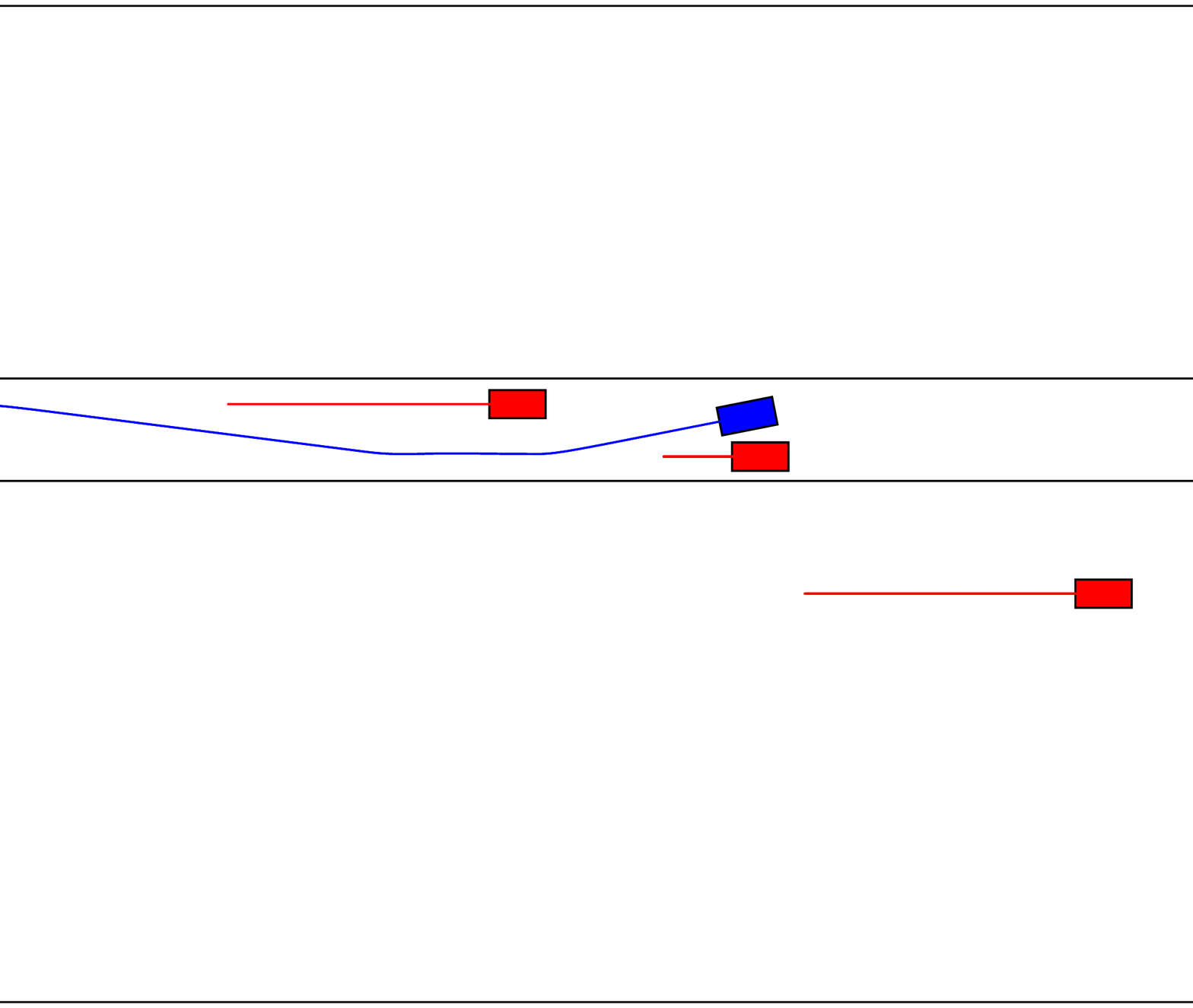}
    \label{lane_3}
   }\vspace{-0.2cm}
   \subfigure[]{
    \includegraphics[width=8.6 cm, height=1.0 cm]{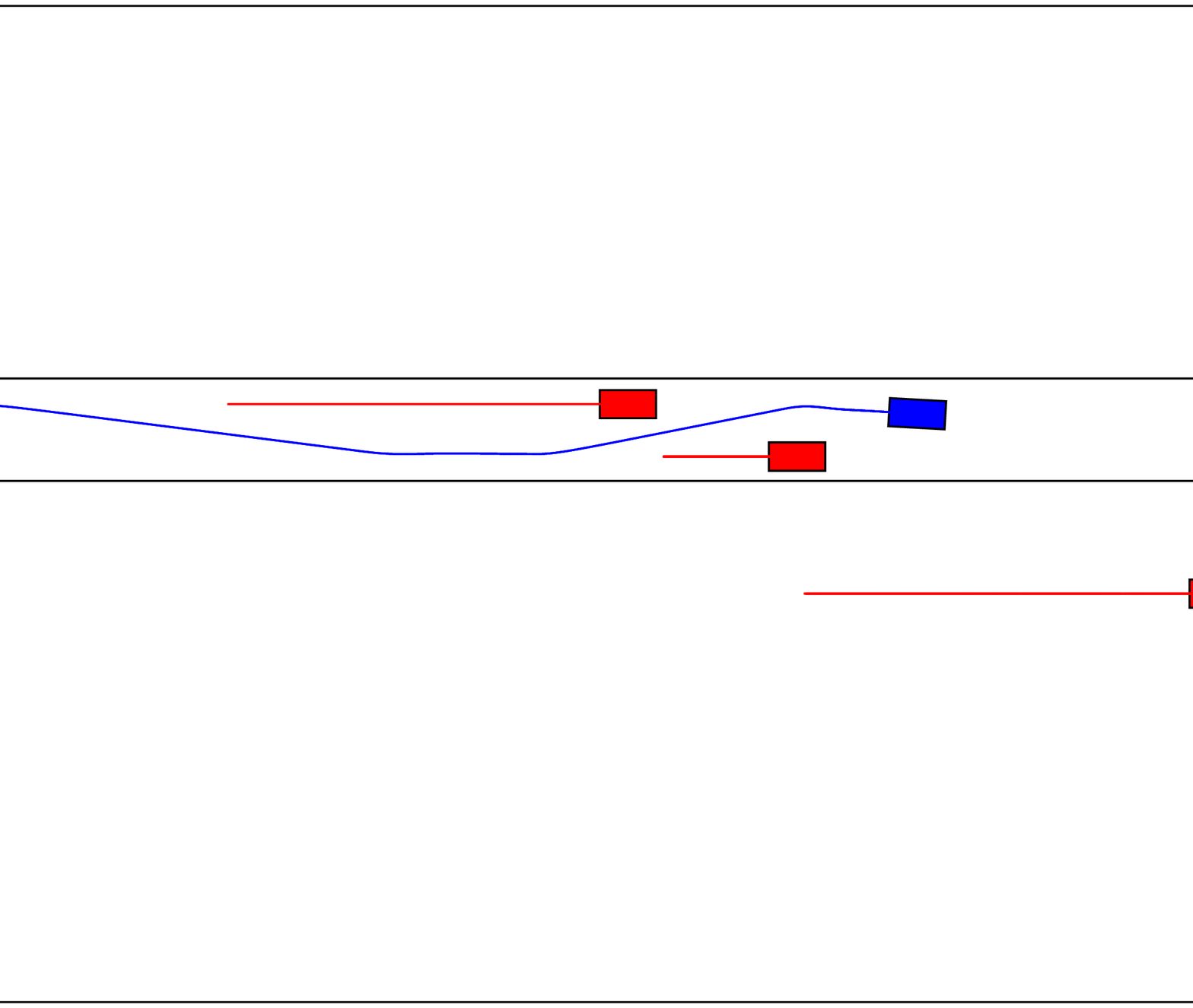}
    \label{lane_4}
   }\vspace{-0.2cm}
   \subfigure[]{
    \includegraphics[width=8.6 cm, height=1.0 cm]{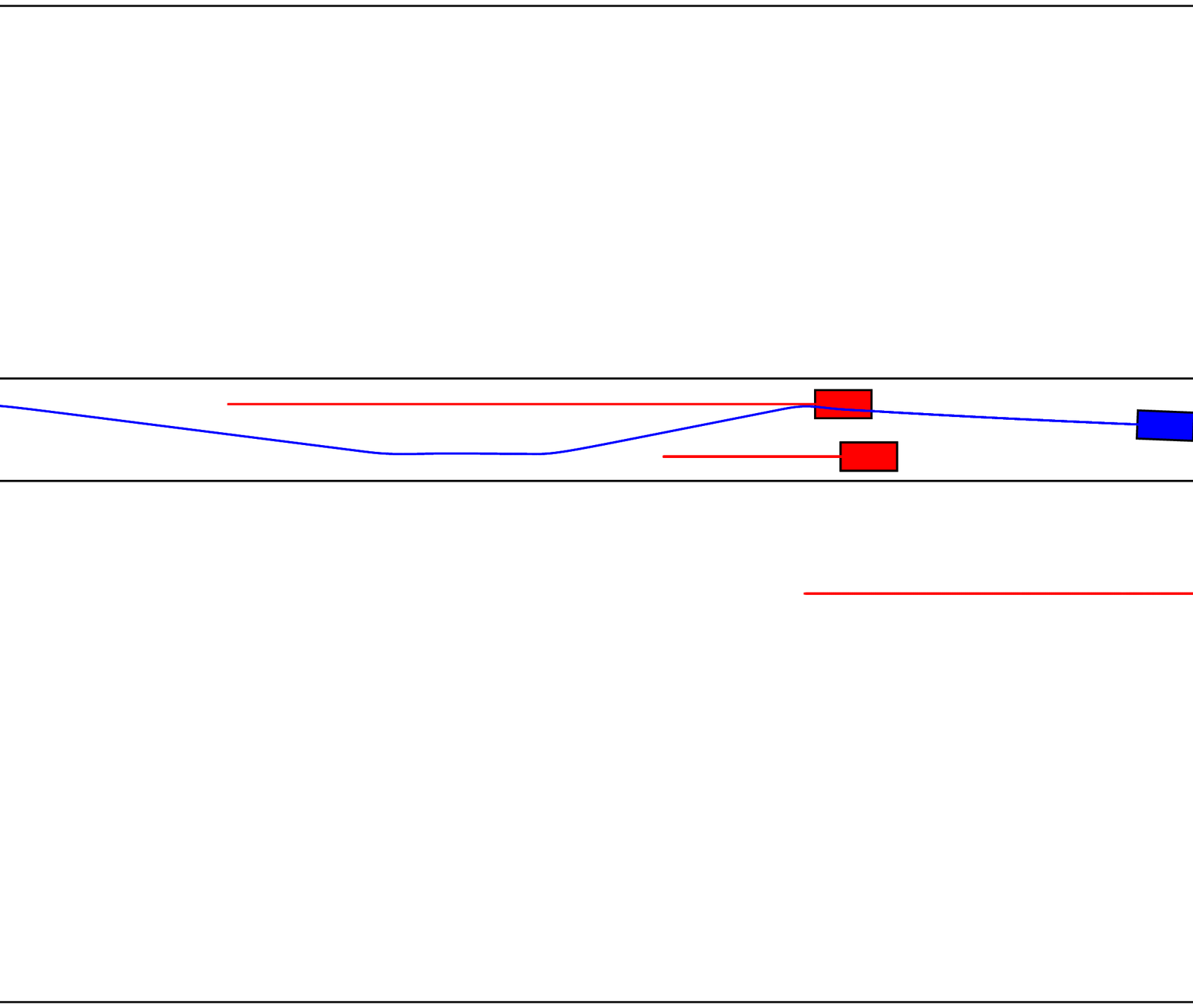}
    \label{lane_5}
   }\vspace{-0.2cm}
   \subfigure[]{
    \includegraphics[width=8.6 cm, height=1.0 cm]{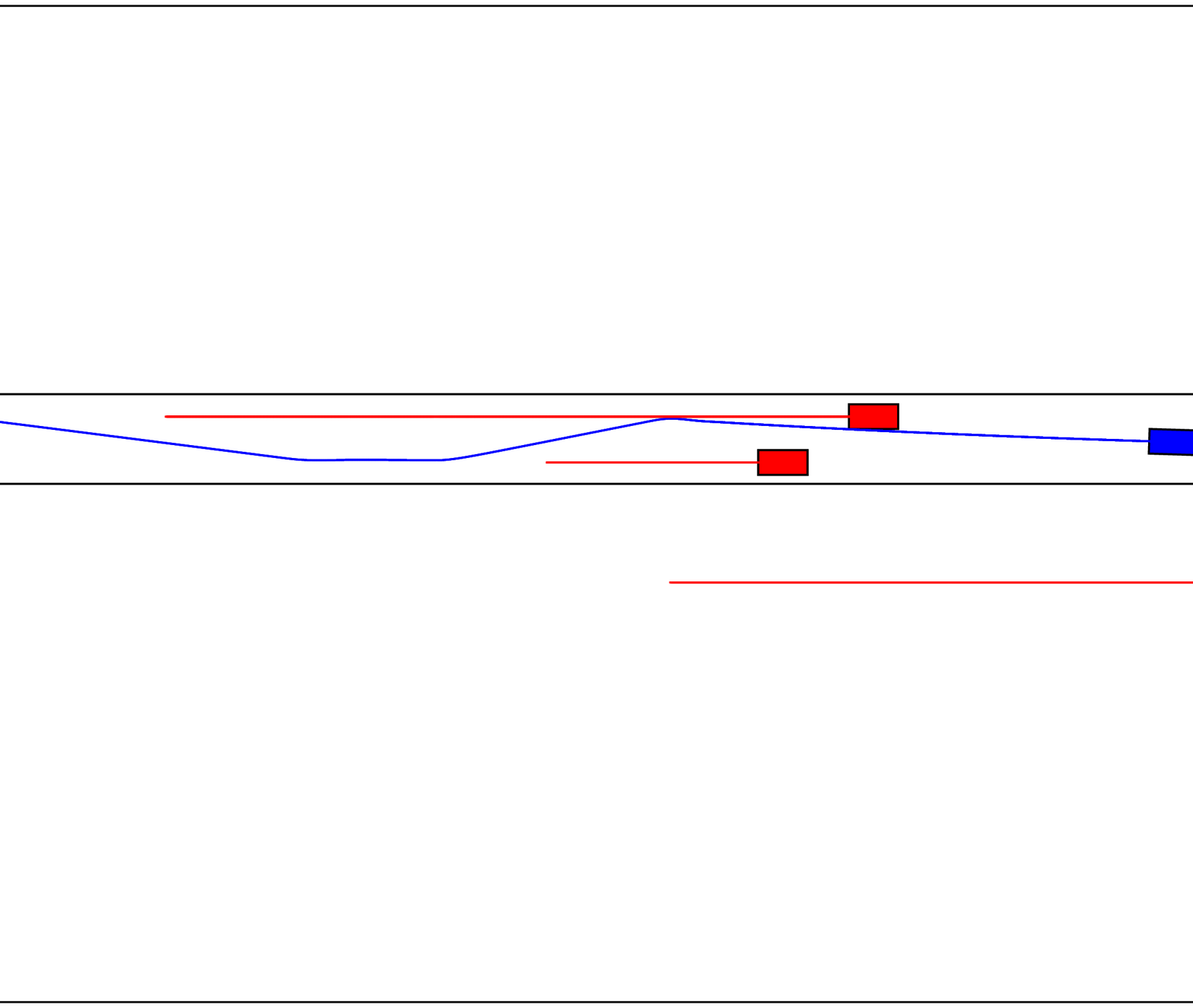}
    \label{lane_6}
   }\vspace{-0.2cm}
   \caption{Lane change by overtaking two dynamic obstacles}
   \end{figure}

\noindent Fig.~\ref{lane_1}-\ref{lane_6} shows the vehicle changing its lane by overtaking two dynamic obstacles ahead. Due to lack of space, we do not show the forward and angular velocity plots but they show similar smoothness as in the previous cases.

\subsection{Computation Time}
\begin{figure}[!h]
\centering
 \includegraphics[width=5.6 cm, height=4.1 cm]{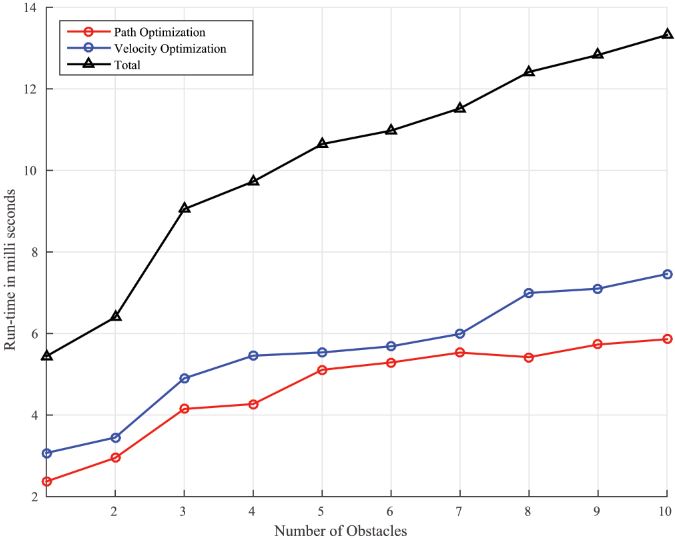}
 \caption{Plot for individual path, forward velocity optimization and total computation time. The values represents the average across all the simulations presented in the previous section.}
 \label{comptime}
\end{figure}

\noindent The mean computation time for each iteration of our MPC loop observed across all the examples presented in the previous section is shown in Fig.~\ref{comptime}. The implementation was done in Python on a 64 bit laptop with  6GB RAM, i5 processor with 2.60 GHz clock speed. As can be seen, the proposed MPC with hierarchical path and velocity optimization can be solved at around $13 ms$ for a scenario with 10 obstacles, resulting in an update rate of almost $77 Hz$.

\section{Conclusion and Future Work}
In this paper, we presented a computationally efficient MPC framework for autonomous driving. The underlying optimization is based on \emph{path velocity decomposition} paradigm and can exactly satisfy the complex kino-dynamic and collision avoidance constraints. In contrast to existing works, the proposed MPC limits its reliance on solving general non-linear optimizations. Instead, the forward velocity optimization layer assumes most of the responsibility of guaranteeing collision avoidance. We have shown that the computation time of our MPC is very low and it  in fact can run at an update rate of about 77 Hz in realistic scenarios.

Our future efforts are focused towards including uncertainty in our MPC framework. A probabilistic variant of TSCC had already been developed in \cite{iros15} and we are looking to extend that formulation to autonomous driving setting.

\bibliographystyle{IEEEtran}  
\bibliography{ref}

\end{document}